\documentclass{bmvc2k}

\title{Examining the Threat Landscape: Foundation Models and Model Stealing}

\newcommand\blfootnote[1]{%
	\begingroup
	\renewcommand\thefootnote{}\footnote{#1}%
	\addtocounter{footnote}{-1}%
	\endgroup
}
	
\addauthor{Ankita Raj}{ankita.raj@cse.iitd.ac.in}{1}
\addauthor{Deepankar Varma$^\dag$}{satwikdpshrit@gmail.com}{2}
\addauthor{Chetan Arora}{chetan@cse.iitd.ac.in}{1}

\addinstitution{
 Indian Institute of Technology Delhi, \\
 India
}
\addinstitution{
Thapar Institute of Engineering and Technology, \\
Patiala, India
}

\runninghead{Raj et al.}{Foundation Models and Model Stealing}

\def\etal{\emph{et al}\bmvaOneDot}

\newcommand{\myfirstpara}[1]{\par \noindent \textbf{#1.}}
\newcommand{\mypara}[1]{\vspace{0.2em} \myfirstpara{#1}}
\newcommand{\myfirstparand}[1]{\par \noindent \textbf{#1}}
\newcommand{\myparand}[1]{\vspace{0.2em} \myfirstparand{#1}}
\DeclareMathOperator*{\argmax}{argmax}

\usepackage{amsmath}
\usepackage{amsfonts}
\usepackage{cleveref}
\usepackage{multirow}
\usepackage{wrapfig}
\usepackage[inline]{enumitem}

\begin{document}

\maketitle

\blfootnote{$^\dag$ Work done as an intern at Indian Institute of Technology Delhi.}


\begin{abstract}
	Foundation models (FMs) for computer vision learn rich and robust representations, enabling their adaptation to task/domain-specific deployments with little to no fine-tuning. However, we posit that the very same strength can make applications based on FMs vulnerable to model stealing attacks. Through empirical analysis, we reveal that models fine-tuned from FMs harbor heightened susceptibility to model stealing, compared to conventional vision architectures like ResNets. We hypothesize that this behavior is due to the comprehensive encoding of visual patterns and features learned by FMs during pre-training, which are accessible to both the attacker and the victim.
	We report that an attacker is able to obtain 94.28\% agreement (matched predictions with victim) for a Vision Transformer based victim model (ViT-L/16) trained on CIFAR-10 dataset, compared to only 73.20\% agreement for a ResNet-18 victim, when using ViT-L/16 as the thief model. 
	We arguably show, for the first time, that utilizing FMs for downstream tasks may not be the best choice for deployment in commercial APIs due to their susceptibility to model theft. We thereby alert model owners towards the associated security risks, and highlight the need for robust security measures to safeguard such models against theft.  Code is available at \url{https://github.com/rajankita/foundation_model_stealing}. 
\end{abstract}


\section{Introduction}
\label{sec:intro}

\begin{figure*}[t]
	\centering
	\includegraphics[width=\textwidth]{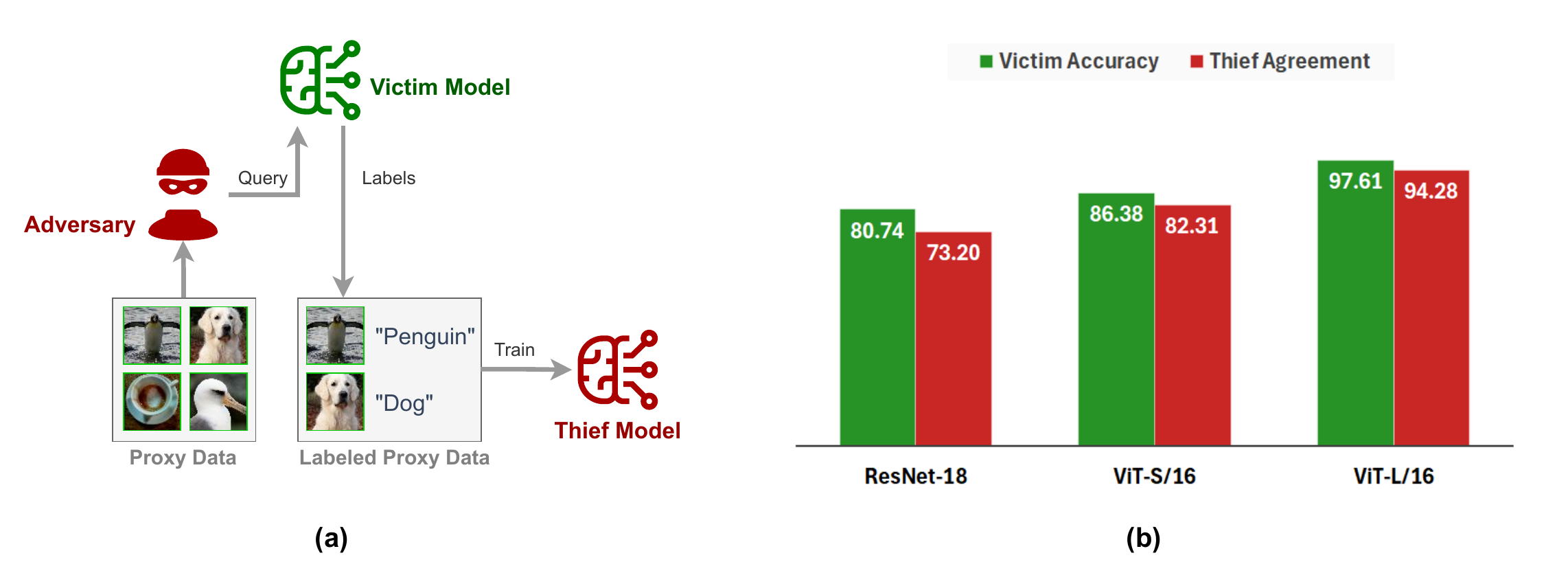}
	\caption{(a) Standard model stealing setup: An adversary picks images from a proxy dataset and queries from the victim model to obtain labels. This labeled proxy dataset is used to train the thief model. (b) Victims derived from foundation models are more prone to stealing: We steal three victim models trained on the CIFAR-10 dataset, using a ViT-L/16 thief. Even though using stronger victims based on foundation models like ViT-L/16 improves victim accuracy, but the agreement between the victim's and thief's predictions also increases at the same time, underlining the increased severity of the threat.}
	\label{teaser}
\end{figure*}

\myfirstpara{Model Stealing Attacks}
Driven by a huge surge in the capabilities of Machine Learning (ML) techniques, many companies now deploy trained ML models on the cloud, and monetize by providing paid access to users via Application Programming Interfaces (APIs). The trained model and the training dataset are often the intellectual property of the company, and therefore the model internals, including the training dataset, model architecture, and weights are kept hidden, providing only black-box access to users. However, such models are vulnerable to model stealing attacks \cite{papernot2017practical,tramer2016stealing,orekondy2019knockoff}, wherein malicious users replicate the behavior of a model by querying the API on a select set of inputs, and training a substitute model on the acquired predictions (see \Cref{teaser}(a)). 
The model thus obtained can either be used as a substitute for the victim model, thereby extracting commercial value from it, or to launch further attacks like adversarial \cite{goodfellow2014explaining,mazeika2022steer} or model inversion \cite{zhang2020secret} on the victim model (called adversarial transfer). In either case, the integrity of the victim model is compromised.

\mypara{Foundation Models}
Recent development in foundation models such as Vision Transformers \cite{dosovitskiy2020image}, and powerful pre-trained encoders like CLIP \cite{radford2021learning} and ALIGN \cite{jia2021scaling} have greatly advanced the field of Computer Vision. 
The availability of foundation models pre-trained on massive datasets has greatly enhanced the capabilities of model owners, who can train task-specific models on downstream applications easily by fine-tuning these models. The resulting downstream models boast high accuracy, causing model owners to be more inclined towards choosing foundation models over conventional vision architectures \cite{he2016deep,inception}. 
While ViTs \cite{dosovitskiy2020image} are known to be more robust compared to convolutional architectures in terms of both adversarial attacks \cite{shao2022adversarial} and common corruptions \cite{paul2022vision}, the model stealing risks associated with downstreaming these models for commercial APIs have not been systematically investigated.

\mypara{Our Focus}
In this paper, we study the vulnerability of image classification APIs based on foundation models to model stealing attacks. The victim models are foundation models (ViTs) fine-tuned on downstream datasets that are made accessible via black-box APIs. We consider both final layer fine-tuning, aka \emph{linear probing}, and \emph{full fine-tuning}. 
Firstly, we show that using more accurate victim models may not always be the best bet for model owners from the perspective of model stealing. This is because, not only the model owners, but attackers can also avail of these powerful pre-trained models. Assuming a well equipped thief, which has access to at least as strong a foundation model as the victim, we show that victim models obtained by fine-tuning from foundation models have a heightened susceptibility to theft compared to smaller models like ResNets. This is due to the extensive knowledge encapsulated within these foundation models, which is now available to both victim and adversary/thief models. 
Secondly, attackers armed with access to a foundation model, by virtue of stronger representation available, can execute model theft with relative ease, when compared to shallower thieves. 

\myparand{Higher Accuracy or Better Privacy?}
By carrying out this study, we aim to raise awareness about the risks associated with using foundation models in the Machine Learning as a Service (MLaaS) setup. While it is always beneficial for an adversary to use foundation models for stealing, the victim needs to be more careful in their choice, and must choose between higher accuracy and better privacy. Our study also underscores the necessity for robust security protocols and countermeasures in the deployment and utilization of these models.

\mypara{Contributions} We make the following contributions:
\begin{enumerate*}[label=\textbf{(\arabic*)}]
	\item We conduct a thorough systematic study on three datasets, seven victim models, and four thief models to evaluate the model stealing vulnerability of victim models obtained from foundation models, particularly ViTs, either via linear probing or by fine-tuning all layers.
	\item Our studies conclude that, under a strong attack wherein the thief has access to pre-trained foundation models, models fine-tuned from ViTs are more vulnerable to theft as compared to models fine-tuned from convolutional architectures. Using a ViT-L/16 thief, we report agreements of 94.28\%, 60.52\% and 62.94\% for victim models based on ViT-L/16 model trained on CIFAR-10, Indoor-67 and Caltech-256 datasets respectively, compared to agreements of 73.20\%, 40.22\% and 46.23\% respectively for ResNet-18 based victims.
	\item Even when the victim model is not fine-tuned from a foundation model, we show that foundation models greatly enhance the attacker's capabilities. When stealing a ResNet-18 victim trained on CIFAR-10 dataset, we report an agreement of 77.12\% for a ViT-B/16 CLIP-based thief, compared to 64.53\% agreement for a ResNet-34 thief (refer \Cref{fig:lp_datasets}).
\end{enumerate*}


\section{Related Work}

\myfirstpara{Model Stealing Attacks}
Model stealing attacks aim to replicate black-box machine learning models, either by extracting exact weights or hyperparameters of the victim model \cite{milli2019model,jagielski2020high,carlini2020cryptanalytic,batina2018csi,hua2018reverse,duddu2018stealing}, or by approximately mimicking the victim model's predictions  \cite{batina2018csi,hua2018reverse,duddu2018stealing}. 
The latter method involves iteratively querying the victim on a designated set of inputs, often referred to as a "proxy" dataset, and subsequently training a substitute model based on the acquired predictions.
Some studies \cite{correia2018copycat,orekondy2019knockoff,pal2020activethief,wang2021black} use natural images from publicly available datasets as the proxy data, devising techniques to select the most informative samples to minimize the attacker's cost. Another school of methods \cite{zhou2020dast, kariyappa2021maze, truong2021data, wang2021delving,sanyal2022towards,Beetham2023} generates synthetic data for querying the victim, eliminating the reliance on natural images but incurring significant costs for the attacker due to the necessity of millions of queries. In this work, we adopt an approximate model stealing setup utilizing natural images, recognizing its practicality and resemblance to real-world attack scenarios.

\mypara{Foundation Models in Computer Vision}
Foundation models, initially conceived within the NLP domain, have found widespread adoption within the computer vision community. Trained on massive amounts of data in a supervised or self-supervised manner, these models learn rich, generalized representations that can serve as the backbone for various downstream tasks. Among these, Vision Transformers (ViTs) \cite{dosovitskiy2020image} stand out as prominent foundation models in vision research that excel in capturing global dependencies and contextual information from visual data using self-attention mechanisms. 
Since the original Vision Transformer (ViT) \cite{dosovitskiy2020image} model, several variants have been proposed including DeiT \cite{touvron2021training}, Swin Transformer \cite{liu2021swin}, DINO \cite{caron2021emerging}, etc. 
Additionally, models like CLIP \cite{radford2021learning} and ALIGN \cite{jia2021scaling} have further expanded the horizon of foundation models in vision, by aligning visual and textual representations into a shared embedding space. 

\mypara{Foundation Models and Model Stealing}
In the NLP domain, Krishna \etal \cite{krishna2019thieves} demonstrated that models fine-tuned from large pretrained language models like BERT can be stolen by issuing semantically irrelevant input queries. In computer vision, model stealing attacks have been studied only for CNN based victim models for the image classification task. Battis \etal evaluated the role of ViTs in model stealing, but only from the attacker's perspective.  An emerging thread of work \cite{liu2022stolenencoder,dziedzic2022difficulty,sha2023can} deals with stealing large pre-trained image encoders like SimCLR \cite{chen2020simple}, MoCo \cite{he2020momentum}, BYOL \cite{grill2020bootstrap} and CLIP \cite{radford2021learning}. These methods aim to steal general-purpose encoders that return rich feature embeddings rather than model posteriors or labels, rendering them highly vulnerable to theft. Differently, in our work, we steal classification models fine-tuned from large foundation models on downstream tasks.


\section{Methodology}

\subsection{Victim Model} 
The victim model is a neural network $f_\textrm{V}$ trained for image classification. The model owner trains $f_\textrm{V}$ on labeled images from training data distribution $P_V(X)$. To ease the training, the model owner uses an open-source pre-trained model, and fine-tunes it on their dataset, called victim dataset. The pre-trained model could either be conventional architectures popularly used in computer vision, e.g., VGG \cite{simonyan2014very} or ResNet \cite{he2016deep}, or modern large foundation models like ViT \cite{dosovitskiy2020image} or CLIP \cite{radford2021learning} which are pretrained on huge datasets and come with strong representation power. The output of the model, $y \in \{1, \dots K\}$, is a distribution over $K$ classes. 

The trained victim model is deployed on a cloud service platform as a black-box. In this setting, the victim's architecture and weights are hidden, but a user can query the model via an API and obtain its predictions on a given image. This setup allows the victim to monetize from their model by charging the user on query basis. 
Several previous works assume the availability of the full probability vector $f_V(x) \in \mathbb{R}^K$ (also known as \emph{soft labels}) as the victim's output. However, many real-world APIs only return the topmost prediction $\argmax_{i\in\{1, \dots, K\}} f_V(x)_i$ for a queried image, also known as \emph{hard-label}. We adapt the hard-label setup in our experiments, on account of it being closest to real-world scenario.

\subsection{Attacker's Goal} 
The attacker's objective is to replicate the behaviour of the victim model. We don't mean to replicate the exact weights of the victim neural network, but to train a substitute/thief model $f_{T}$ that is functionally equivalent to the victim model in terms of predictions on the victim's held-out test set. 

\subsection{Attack Method} \label{sec:attack_method}
A major constraint for the attacker is that it does not have access to the victim's training data distribution $P_V(X)$. It therefore, uses a \textit{proxy} distribution $P_A(X)$ to query the victim model. The attacker selects a subset of images $\{x^i\}_{i=1}^m$ from $P_A(X)$ and receives predictions for the queried images, thus constructing a labeled set $\mathcal{D}_{l}=\{(x^i, f_\textrm{V}(x^i)\}_{i=1}^{m}$ of size $m$ which is then used to train the thief model $f_{T}$. Fundamentally, a thief model uses information obtained from the victim in the form of queried labels to learn similar decision boundaries as the victim model. The performance of the thief model depends on the following factors:

\mypara{Proxy dataset}
For computer vision applications, the proxy dataset can be constructed from publicly accessible natural images, or by generating synthetic images. In line with previous works \cite{orekondy2019knockoff,pal2020activethief}, we use large scale publicly available datasets of natural images as the proxy distribution.

\mypara{Query selection method} For practical model stealing attacks, the thief has to work under limited query budgets. As such, there is a large body of works dedicated to selecting the best set of samples from the proxy dataset to query the victim model. This includes the reinforcement-learning methods by \cite{orekondy2019knockoff}, active learning based methods by \cite{pal2020activethief} and GRAD-CAM based methods by \cite{wang2021black}. In this work, we adopt the simple yet effective Random selection strategy \cite{orekondy2019knockoff} for most of our experiments. The impact of changing the sample selection technique is studied in \Cref{sec:ablation}.

\mypara{Thief architecture}
Typically, the attacker has no knowledge of the victim model's architecture or hyper-parameters. However, several works on model stealing assume that the attacker uses the same architecture as the victim model, for ease of experimentation, and also owing to the belief that the thief model's architecture does not significantly affect the model stealing performance. However, given the easy availability of large and powerful foundation models, it would be prudent for an attacker to use these larger models instead. We therefore assume that an informed thief is able to use open-source pretrained models that are available on the internet, and fine-tunes the model on the labeled dataset $\mathcal{D}_{l}$ constructed from the proxy data. For completeness, we also study the special case scenario when the thief model has the same architecture as the victim (in the Supplementary).


\section{Experimental Setup}

\subsection{Victim Datasets and Architectures}

\mypara{Datasets} 
We adopt three commonly-used image recognition datasets to train our victim models.
 \textbf{CIFAR-10} \cite{krizhevsky2009learning}, \textbf{Caltech-256} \cite{griffin2007caltech}, and  \textbf{Indoor-67} \cite{quattoni2009recognizing}. Of these, CIFAR-10 and Caltech-256 are general-purpose object recognition datasets with 10 and 256 object classes respectively, while Indoor-67 is a fine-grained classification dataset comprising 67 types of indoor scenes.

\mypara{Architectures}
We work with seven different victim architectures in our experiments. These include four conventional CNN architectures: \textbf{ResNet-18}, \textbf{ResNet-34}, \textbf{ResNet-50} and \textbf{ResNet-101} \cite{he2016deep} pre-trained on ImageNet-1K dataset \cite{russakovsky2015imagenet} containing 1.2M images, and three vision foundational models: \textbf{ViT-S/16}, \textbf{ViT-B/16} and \textbf{ViT-L/16} \cite{dosovitskiy2020image} pre-trained on the larger ImageNet-21K dataset \cite{deng2009imagenet} containing 14M images from 21,841 classes. 
The power of these models comes not only from the deeper architectures, but also from stronger pre-training on massive datasets.  

\mypara{Training details}
Fine-tuning on ImageNet-pretrained weights is a widely adopted method of training CNN models. However, with the advent of powerful foundation models, fine-tuning only the final layer (a.k.a linear probing) is gaining popularity, as it offers good results at low training costs. We, therefore, consider both scenarios in our experiments. 
All models are initialized using publicly available ImageNet \cite{russakovsky2015imagenet} pre-trained weights: the ResNet models are pre-trained on ImageNet-1K, and the ViTs are pre-trained using the larger ImageNet-21K dataset. All pre-trained weights are obtained from open-source libraries like Pytorch hub \cite{torchhub} or timm \cite{rw2019timm}.
All victim models are trained using SGD optimizer for 100 epochs, with a momentum of 0.9, weight decay of $5 \times 10^{-4}$, and batch size of 128. An initial learning rate is $0.002$ is decayed by a factor of $10$ after every 30 epochs. All input images are resized to $224\times224$ and augmented using random cropping and random horizontal flipping.

\subsection{Thief Datasets and Architectures} 

\mypara{Dataset}
We use publicly available unannotated images from the training set of ILSVRC-2012 challenge dataset \cite{russakovsky2015imagenet} as the thief's proxy dataset. Commonly known as ImageNet, the dataset contains 1.2 million images of varying sizes, of which we only use a subset of 128K images resized to $224\times224$. 

\mypara{Architectures}
We consider multiple model architectures for the thief model. These include a CNN model ResNet-34 \cite{he2016deep}, two ViT models: ViT-B/16 and ViT-L/16 \cite{dosovitskiy2020image}, as well as a multimodal foundation model ViT-B/16 CLIP \cite{radford2021learning}. 
In general, we assume that the thief has access to all publicly available pre-trained models that the victim has access to.  

\mypara{Training details}
Thief models are trained using SGD optimizer for 100 epochs, with a momentum of 0.9, weight decay of $5 \times 10 ^{-4}$, and a batch size of 128 (for ResNet-34), 32 (for ViT-B/16 and ViT-B/16 CLIP) or 64 (ViT-L/16). The initial learning rate is $0.001$, and it is decayed by a factor of $10$ after 50 epochs. 
For stealing CIFAR-10 vicitm, input images are resized to $224\times224$, and augmented using random cropping, random rotation and random horizontal flip. For stealing other victims, we use random crop, random horizontal flip and RandAugment \cite{cubuk2020randaugment}. 
All networks are initialized from ImageNet-pretrained weights. For ResNet thieves, all layers are fine-tuned, whereas for ViT's only the final layer is fine-tuned.
\mypara{Sample Selection} For majority of our experiments, we consider the random-adversary based sample selection method from Knockoff Nets \cite{orekondy2019knockoff}, and limit the thief's query budget to 5000 samples. Of these, $10\%$ samples are set aside as validation data for hyper-parameter selection, and the rest are used for training. 
In \Cref{sec:ablation}, we evaluate the impact of the sample selection method and query budget on our findings. 

\subsection{Evaluation Metrics} 
We use two evaluation metrics prevalent in the literature. \emph{Accuracy} measures the performance of the thief model on the victim model's held-out test set, while \emph{Agreement} measures how often the thief's prediction matches the victim's. Note that the held-out victim dataset is not accessible to the thief, and is only used for evaluation.


\section{Experiments and Results}

\begin{wrapfigure}{R}{0.45\textwidth}
	\centering
	\includegraphics[width=0.44\textwidth]{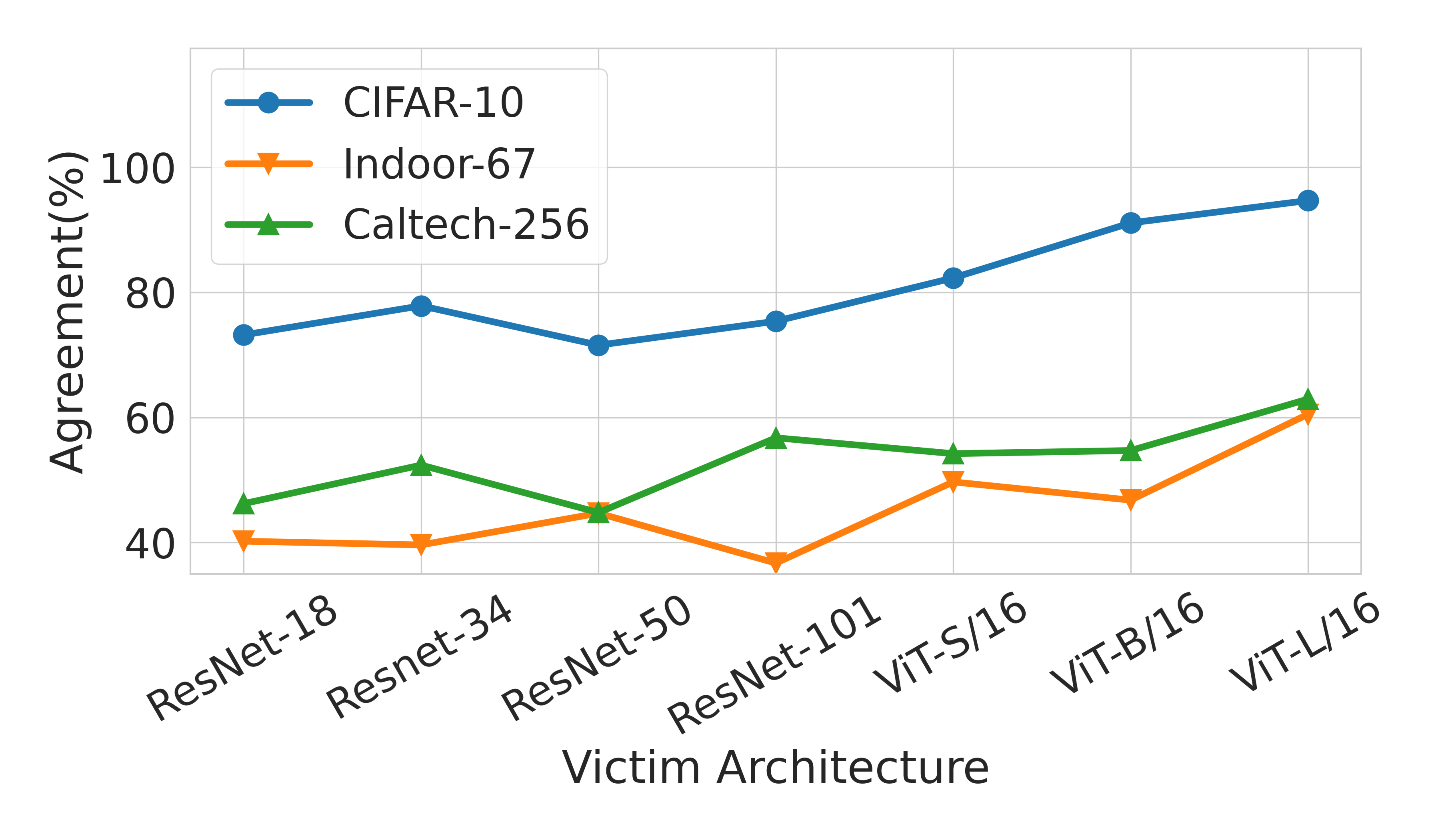}
	\vspace{-3mm}
	\caption{
		\small{Agreement between multiple linear-probed victims (trained on three datasets) and a linear-probed ViT-L/16 thief.}
	}
	\label{fig:expt1_agr}
	\vspace{-2mm}
\end{wrapfigure}

\subsection{Foundation Model Victims are Easy to Steal} \label{sec:expt1}
In this section, we try to answer the question: \emph{Is a victim fine-tuned from a foundation model more vulnerable to model stealing compared to a victim fine-tuned from a shallower model?} 
To do so, we train multiple victim models, including both conventional architectures and foundation models. For the thief model, we assume a well-equipped attacker who utilizes large, publicly available pre-trained foundation models, as reasoned in \ref{sec:attack_method}. 

\mypara{Linear-probed Victims} \label{main_expt}
We first consider victim models that are trained by fine-tuning the last layer of the respective pre-trained models, which is a common practice with foundation models. The thief model is a ViT-L/16, which is also linear-probed. We report the victim models' accuracy, along with the accuracy and agreement of the thief models in \Cref{tab:lp}. We draw the following observations.
Firstly, using foundation models leads to an increase in victim model accuracy, which is expected.
Secondly, the ViT-L/16 thief's accuracy is higher for victims derived from foundation models (ViT-S/16, B/16 and L/16). This can be attributed partly to the higher victim model accuracy for foundation models. Therefore, we also report the agreement between the victim and thief models, which directly quantifies the similarity between the predictions of the victim and thief.
Thirdly, we see that agreements are also higher for ViT victims compared to ResNets (see \Cref{fig:expt1_agr}). On the Indoor-67 dataset for instance, agreement for a ResNet-18 victim is only $40.22\%$, as compared to $60.52\%$ for a ViT-L/16 victim. These findings suggest that while foundation models like ViTs offer higher accuracy for the victim, they are also more susceptible to theft compared to non-foundation models, particularly when targeted by well-equipped attackers who utilize foundation models themselves.

\newcommand{\xx}{{$\times$}}
{\renewcommand{\arraystretch}{1.1} 
\begin{table}[t]
	\begin{center}
		\resizebox{0.99\linewidth}{!}{
			\begin{tabular}{ll|ccccccc}
				\hline
				& Victim architecture     & ResNet-18 & ResNet-34 & ResNet-50 & ResNet-101 & ViT-S/16 & ViT-B/16 & ViT-L/16 \\
				& Params (M)         & 11.18     & 21.29     & 23.53  & 42.52  & 21.66   & 85.80  & 303.31 \\
				& Pretraining dataset & IN-1K & IN-1K & IN-1K & IN-1K & IN-21K & IN-21K & IN-21K \\
				\hline
				\multirow{3}{*}{\rotatebox{90}{CIFAR}} 
				& Victim accuracy & 80.74 & 81.96 & 80.24 & 84.64 & 86.38 & 94.12 & 97.61 \\
				& Thief accuracy  & 82.97 & 88.92 & 81.17 & 83.92 & 90.49 & 93.79 & 94.53\\
				& Thief agreement & 73.20 & 77.85 & 71.56 & 75.39 & 82.31 & 91.10 & 94.28 \\
				\hline
				\multirow{3}{*}{\rotatebox{90}{Indoor}}
				& Victim accuracy & 54.70 & 58.06 & 45.97 & 51.49 & 78.51 & 82.76  & 87.39  \\
				& Thief accuracy  & 36.87 & 39.70 & 34.10 & 29.85 & 51.49 & 46.57 & 59.03 \\
				& Thief agreement & 40.22 & 39.63 & 44.70 & 36.72 & 49.70 & 46.79 & 60.52   \\
				\hline
				\multirow{3}{*}{\rotatebox{90}{Caltech}}
				& Victim accuracy & 67.75 & 74.36 & 58.47 & 74.78 & 84.78 & 87.67 & 94.13 \\
				& Thief accuracy  & 47.42 & 54.86 & 35.38 & 53.63 & 58.20 & 57.91 & 62.83 \\
				& Thief agreement & 46.23 & 52.39 & 44.80 & 56.75 & 54.22 & 54.72 & 62.94 \\
				\hline
			\end{tabular}
		}
	\end{center}
	\caption{Model stealing results for \textbf{linear-probed victim} models on three victim datasets. Thief model is ViT-L/16. IN stands for ImageNet.}
	\label{tab:lp}
\end{table}

\begin{table}[t]
	\begin{center}
		\resizebox{0.99\linewidth}{!}{
			\begin{tabular}{ll|cccccc}
				\hline
				& Victim architecture     & ResNet-18 & ResNet-34 & ResNet-50 & ResNet-101 & ViT-S/16 & ViT-B/16 \\
				& Params (M)         & 11.18     & 21.29     & 23.53  & 42.52  & 21.66   & 85.80 \\
				& Pretraining dataset & IN-1K & IN-1K & IN-1K & IN-1K & IN-21K & IN-21K \\
				\hline
				\multirow{3}{*}{\rotatebox{90}{CIFAR}} 
				& Victim accuracy & 96.54 & 97.41 & 97.62 & 98.42 & 98.33 & 98.33 \\
				& Thief accuracy  & 92.17 & 94.74 & 94.74 & 93.80 & 94.67 & 91.48 \\
				& Thief agreement & 90.89 & 93.96 & 93.80 & 93.10 & 94.01 & 91.06 \\
				\hline
				\multirow{3}{*}{\rotatebox{90}{Indoor}}
				& Victim accuracy & 75.60 & 78.73 & 81.64 & 82.61 & 87.01 & 86.19 \\
				& Thief accuracy  & 43.21 & 41.64 & 47.16 & 49.63 & 55.52 & 52.83 \\
				& Thief agreement & 41.27 & 41.42 & 48.28 & 49.48 & 55.97 & 53.96  \\
				\hline
				\multirow{3}{*}{\rotatebox{90}{Caltech}}
				& Victim accuracy & 76.78 & 81.39 & 86.00 & 88.70 & 93.44 & 95.14 \\
				& Thief accuracy  & 47.73 & 56.14 & 61.77 & 63.20 & 63.00 & 57.57 \\
				& Thief agreement & 44.58 & 52.78 & 60.23 & 61.67 & 62.25 & 56.98 \\
				\hline
			\end{tabular}
		}
	\end{center}
	\caption{Model stealing results for \textbf{fully fine-tuned victim} models on three different victim datasets. Thief model is ViT-L/16. IN stands for ImageNet.}
	\label{tab:fft}
\end{table}

\mypara{Fully Fine-tuned Victims} 
In this section, we evaluate victims obtained by fine-tuning all layers of the backbone models, while the thief is still a linear-probed ViT-L/16. We show the results for stealing multiple victim models using a ViT-L/16 thief in \Cref{tab:fft}, and once again, observe higher thief agreements for ViT victims compared to ResNets.

\begin{figure}[t]
	\centering
	\includegraphics[width=\textwidth]{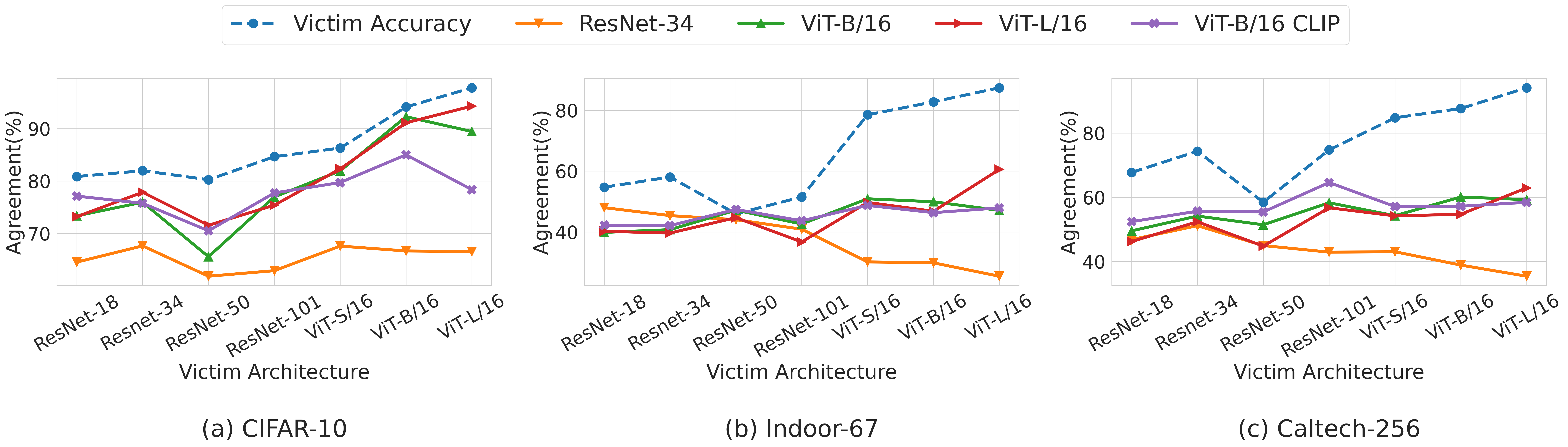}
	\caption{Agreement for thieves based on foundation models (ViTs) \textit{vs.} a ResNet-34 based thief. Victim models are \textbf{linear-probed}. }
	\label{fig:lp_datasets}
\end{figure}

\subsection{Foundation Models make Strong Thieves}
So far, we have analyzed the impact of using foundation models on the victim's end. In this section, we look at the other perspective and try to answer the question: \emph{How do foundation models affect the capabilities of a thief?} 
We vary the thief's model architecture and study its impact on model stealing performance for various victim models. We compare the agreements for four thief architectures: ResNet-34, ViT-B/16, and ViT-L/16, and ViT-B/16 CLIP, for linear-probed victims in \Cref{fig:lp_datasets} and fully fine-tuned victims in \Cref{fig:fft_datasets}. We can see that for a given victim architecture, especially deeper architectures, a foundation model thief achieves higher agreement as compared to a regular ResNet thief. More importantly, the gap in performance between the ResNet thief and ViT thieves increases, as victim models become deeper.
When stealing smaller capacity victim models, the ResNet thief performs as well as, or even better than the stronger ViT thieves, especially for Indoor-67 and Caltech-256 datasets. But when the victim model itself is a higher capacity model, the ResNet thief manages to recover only a small proportion of the victim model's accuracy. For instance, in \Cref{fig:lp_datasets}, when stealing a ViT-B/16 victim trained on Indoor-67 dataset, a ResNet-34 thief achieves an agreement of only 25.44\%, whereas a ViT-L/16 thief achieves 46.79\% agreement. Overall, we observe that foundation models serve as better thieves, particularly when the victims are also derived from foundation models.

\begin{figure}[t]
	\centering
	\includegraphics[width=\textwidth]{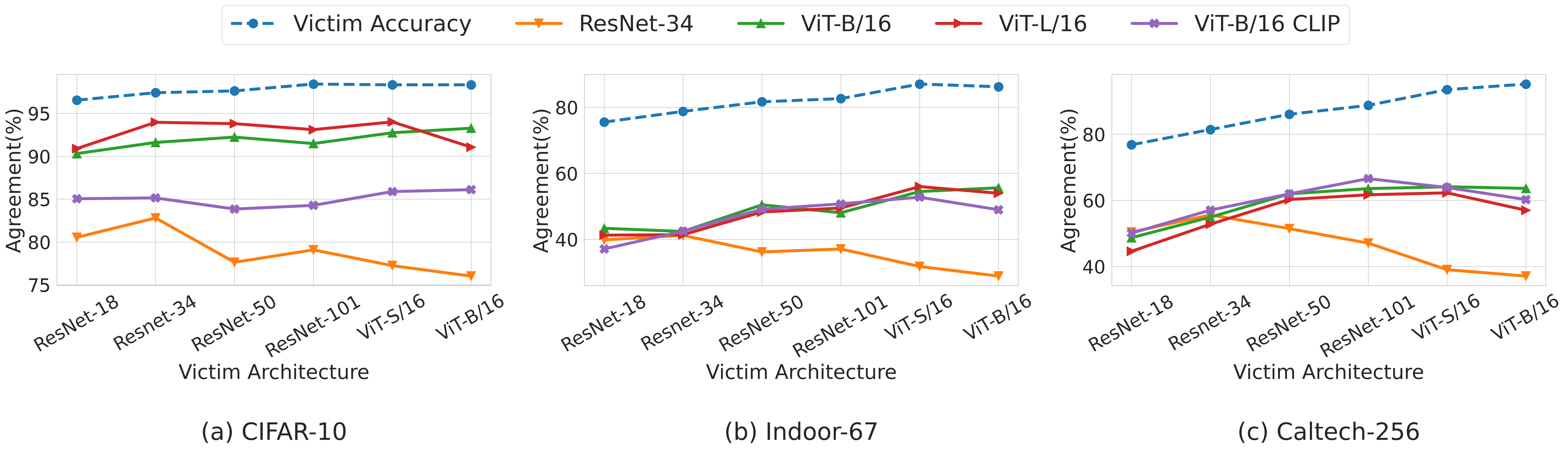}
	\caption{Agreement for thieves based on foundation models (ViTs) \textit{vs.} a ResNet-34 based thief. Victim models are \textbf{fully fine-tuned}. }
	\label{fig:fft_datasets}
\end{figure}

\subsection{Qualitative Analysis} \label{sec:tsne}
To understand why stealing victims fine-tuned from foundation models is more successful, we visualize embeddings of the pretrained backbone models and the victim models in \Cref{fig:tsne_plots}. Using the CIFAR-10 dataset for ease of visualization, we select five victim models trained by linear probing. For the backbone, we plot t-SNE embeddings \cite{van2008visualizing} of the penultimate layer on CIFAR-10 test set. Since the victim models are obtained by fine-tuning only the classification head, we do not expect a change in embeddings of the penultimate layer. We therefore, plot the final classification layer embeddings for the victim models. Notably, for foundation models like ViT-B/16 and ViT-L/16, the backbone is already strong enough to form well-separated clusters on CIFAR-10 despite not having been trained on CIFAR-10. This clear separation of classes extends to the corresponding victim models. We hypothesize that due to the rich representations captured by the backbone architecture in foundation models, the primary computational burden in the victim model resides within this backbone structure, with relatively minimal reliance on the classification head. Consequently, when the thief model possesses a similarly robust backbone architecture, stealing the classification head becomes significantly easier. This stands in contrast to architectures such as ResNet-18, where the pre-trained features within the backbone are comparatively less potent, thereby rendering the task of stealing the classification head more challenging.

\begin{figure*}[t]
	\centering
	\includegraphics[width=\textwidth]{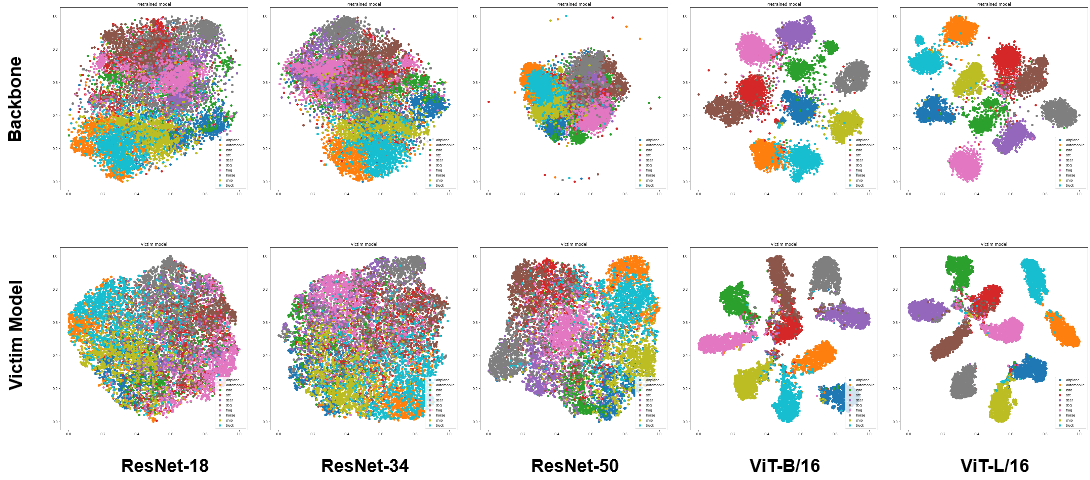}	
	\caption{t-SNE  visualizations {\protect\cite{van2008visualizing}} of embeddings for backbone models (top row), and corresponding victim models (bottom row) trained on CIFAR-10 dataset using linear probing method. Observe that the clusters from the backbone models are much more well-separated for the foundation models (ViT-B/16 and ViT-L/16) compared to ResNets. }
	\label{fig:tsne_plots}
\end{figure*}

\subsection{Ablation Study} \label{sec:ablation}

\mypara{Impact of query budget} We vary the query budget in the range \{2K, 5K, 10K, 20K\} and compute agreement for different linear-probed victims on the Indoor-67 dataset, and a ViT-B/16 thief. Our findings indicate that the same trend holds for all query budgets: foundation models can be stolen with higher agreement compared to ResNets. While the gap in performance of ResNets and ViTs is less for lower budgets, increasing the number of queries proves more detrimental for ViTs than for ResNets (figure in Supplementary). 

\mypara{Impact of sample selection method} So far, we had used Knockoff Nets' random sample selection method \cite{orekondy2019knockoff} to select query points from the proxy dataset. We deploy two more query-set selection strategies: Entropy-based and kCenter-based methods from ActiveThief \cite{pal2020activethief} and observe similar trends in agreement for foundation models and ResNets, for linear-probed victims on the Indoor-67 dataset, stolen using a ViT-B/16 thief (figure in Supplementary), proving the generality of our findings.


\section{Conclusion}
We studied the susceptibility of image classification models fine-tuned from powerful foundation models to model stealing attacks. Our findings reveal that victim models derived from foundation models exhibit greater vulnerability to such attacks from strong, foundation-model based thieves, compared to those derived from shallower backbones. In stark contrast to the celebrated robustness of foundation models against adversarial and natural corruptions, our study sheds light on their heightened susceptibility in the model stealing context. This highlights a crucial trade-off between privacy and accuracy inherent in deploying models fine-tuned from foundation models in commercial APIs, and emphasizes the necessity for enhanced security measures in model deployment strategies.

\section*{Acknowledgement} 
We acknowledge and thank the funding support from AIIMS Delhi-IIT Delhi Center of Excellence in AI funded by Ministry of Education, government of India, Central Project Management Unit, IIT Jammu with sanction number IITJMU/CPMU-AI/2024/0002. 

\bibliography{msa_ref}

\clearpage
\setcounter{table}{0}
\renewcommand{\thetable}{S\arabic{table}}%
\setcounter{figure}{0}
\renewcommand{\thefigure}{S\arabic{figure}}%
\begin{center} 
	{\Large Supplementary Material }
\end{center}

\appendix

\section{Dataset Details}

\begin{wrapfigure}{R}{0.55\textwidth}
	\centering
	\includegraphics[width=\linewidth]{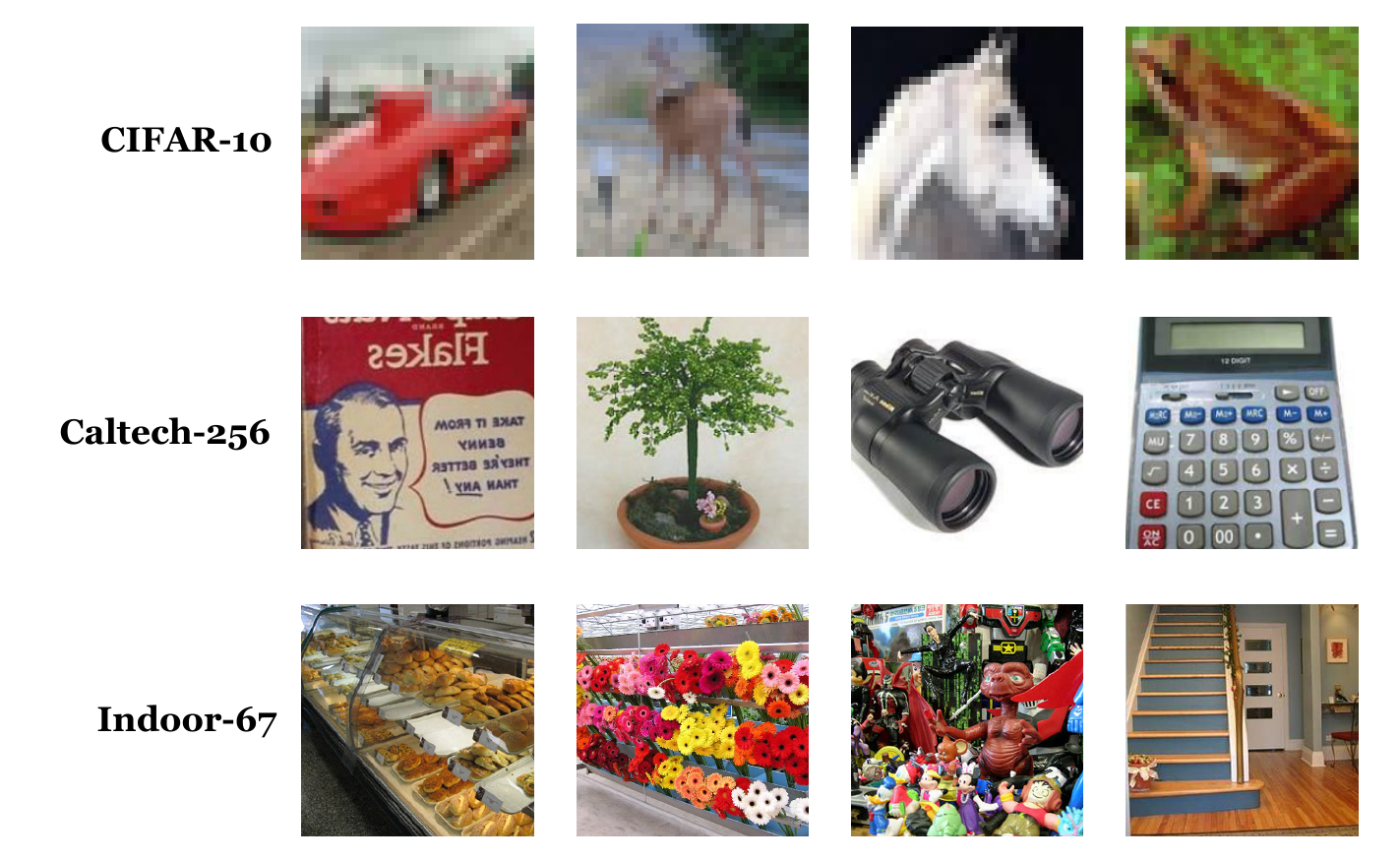}
	\caption{Sample images from victim datasets.}
	\label{fig:sample_images}
\end{wrapfigure}

The CIFAR-10 dataset contains images of resolution $32\times32$. The training set contains $50,000$ images ($5,000$ from each class), and the test set contains $10,000$ images.
The Indoor-67 dataset contains $15,620$ images, with a minimum resolution of 200 pixels in the smaller axis. The dataset is split into $14,280$ training and $1,340$ test images.
Caltech-256 contains images of varying sizes spanning 256 object classes. Since no official train-test split has been reported, we keep aside 25 images from each class for testing, resulting in a training set of $23,380$ images and test set of $6,400$ images.
\Cref{fig:sample_images} shows sample images from each of the three victim datasets. 

\section{Additional Results}

\subsection{Same Victim and Thief Architectures} \label{sec:same_victim_thief}
We had assumed so far that the thief does not have knowledge of the victim model's architecture, and therefore uses a large pre-trained model that is publicly available. Another scenario usually considered in the  model stealing literature \cite{orekondy2019knockoff,pal2020activethief} is when both victim and thief model are based on the same neural network architecture. We study this scenario in this section, where the thief has knowledge of the victim's architecture, and therefore, fine-tunes from the same pre-trained model as the victim. We report the accuracies of the victim and thief models, along with the agreement between victim and thief in \Cref{tab:victim=thief}, and plot the agreements in \Cref{fig:victim_equals_thief} for better visualization. In addition to the seven victim models considered in \Cref{sec:expt1}, we include two self-supervised foundation models: DINO \cite{caron2021emerging} and CLIP \cite{radford2021learning}, pre-trained on the ImageNet-1K and LAION-2B datasets respectively. 

We observe that for the high capacity ViT models with stronger pre-training, both accuracy and agreement of the thief models increase. Moreover, self-supervised foundation models (CLIP and DINO) are also more vulnerable to model stealing compared to ResNets.  This reiterates our finding from \Cref{sec:expt1} that even though stronger foundation models increase the accuracy of victim models, it increases the risk of model stealing too.

\begin{table}[t]
	\begin{center}
		\resizebox{0.99\linewidth}{!}{
			\begin{tabular}{ll|ccccccccc}
				\hline
				& Victim arch     & ResNet-18 & ResNet-34 & ResNet-50 & ResNet-101 & ViT-S/16 & ViT-B/16 & ViT-B/16 DINO & ViT-B/16 CLIP & ViT-L/16 \\
				& Params (M)         & 11.18     & 21.29     & 23.53  & 42.52  & 21.66   & 85.80  & 85.80 & 85.80 & 303.31 \\
				& Pretraining dataset & IN-1K & IN-1K & IN-1K & IN-1K & IN-21K & IN-21K & IN-1K & LAION-2B & IN-21K \\
				\hline
				\multirow{3}{*}{\rotatebox{90}{CIFAR}} 
				& Victim accuracy & 80.74 & 81.96 & 80.24 & 84.64 & 86.38 & 86.93 & 95.06 & 96.26 & 97.61 \\
				& Thief accuracy  & 67.02 & 72.12 & 68.15 & 77.97 & 90.70 & 86.42 & 79.63 & 89.79 & 96.55 \\
				& Thief agreement & 69.42 & 72.42 & 65.53 & 75.51 & 84.63 & 80.96 & 80.96 & 90.74 & 94.68 \\
				\hline
				\multirow{3}{*}{\rotatebox{90}{Indoor}}
				& Victim accuracy & 54.70 & 58.06 & 45.97 & 51.49 & 78.51 & 82.76 & 80.22 & 83.58 & 87.39 \\
				& Thief accuracy  & 28.06 & 31.27 & 22.09 & 26.04 & 57.46 & 49.70 & 52.84 & 63.58 & 57.39 \\
				& Thief agreement & 41.94 & 45.22 & 42.69 & 39.79 & 58.73 & 51.49 & 57.91 & 69.70 & 58.51 \\
				\hline
				\multirow{3}{*}{\rotatebox{90}{Caltech}}
				& Victim accuracy & 67.75 & 74.36 & 58.47 & 74.78 & 84.78 & 87.67 & 87.66 & 91.98 & 94.13 \\
				& Thief accuracy  & 18.83 & 37.17 & 14.88 & 28.61 & 63.02 & 63.23 & 59.78 & 68.30 & 61.30 \\
				& Thief agreement & 25.27 & 44.89 & 34.56 & 38.25 & 61.23 & 61.14 & 62.19 & 71.39 & 60.95 \\
				\hline
			\end{tabular}
		}
	\end{center}
	\caption{Model stealing results when thief model architecture is same as victim model architecture. Both victim and thief models are linear probed. IN stands for ImageNet.}
	\label{tab:victim=thief}
\end{table}

\begin{figure}[t]
	\centering
	\includegraphics[width=0.7\linewidth]{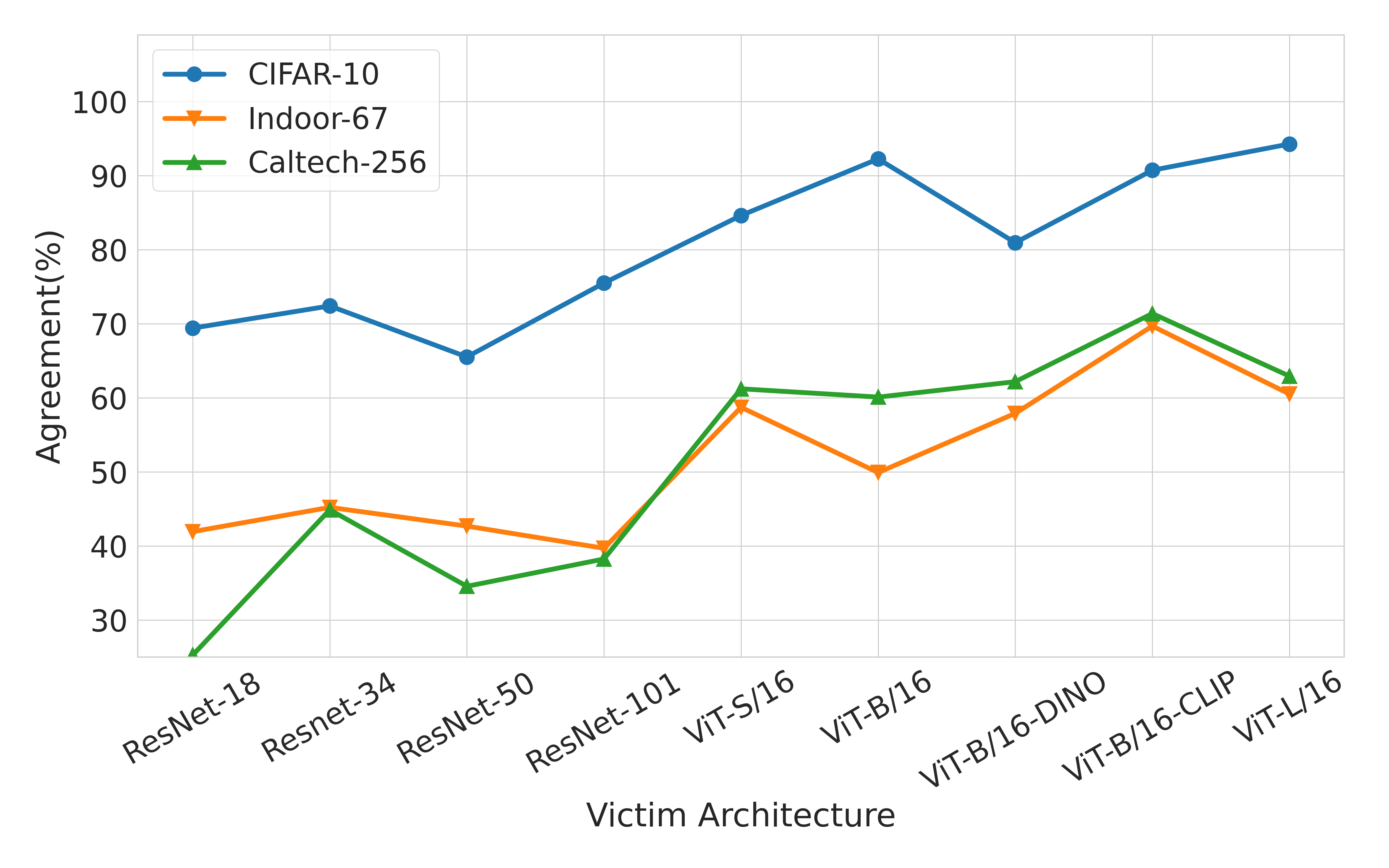}	
	\caption{Agreement between victim and thief models. Both victim and thief share the same architecture.}
	\label{fig:victim_equals_thief}
\end{figure}

\section{Extensions of Existing Results} 

\subsection{Ablation study from Main Paper} We provide here detailed results to support the ablation studies reported in \Cref{sec:ablation}. \Cref{fig:ablation1} and \Cref{fig:ablation2} show the impact of varying the query budget and sample selection technique respectively, for victims trained on the Indoor-67 dataset using the linear probing method. The thief is a ViT-B/16 model (chosen because of faster training time) and is trained using linear probing. We observe a similar trend in both cases: the ViT models with their rich feature representations are stolen with higher agreements compared to ResNet models. 

\begin{figure}[t]
	\centering
	\begin{minipage}{.47\textwidth}
		\centering
		\includegraphics[width=\textwidth]{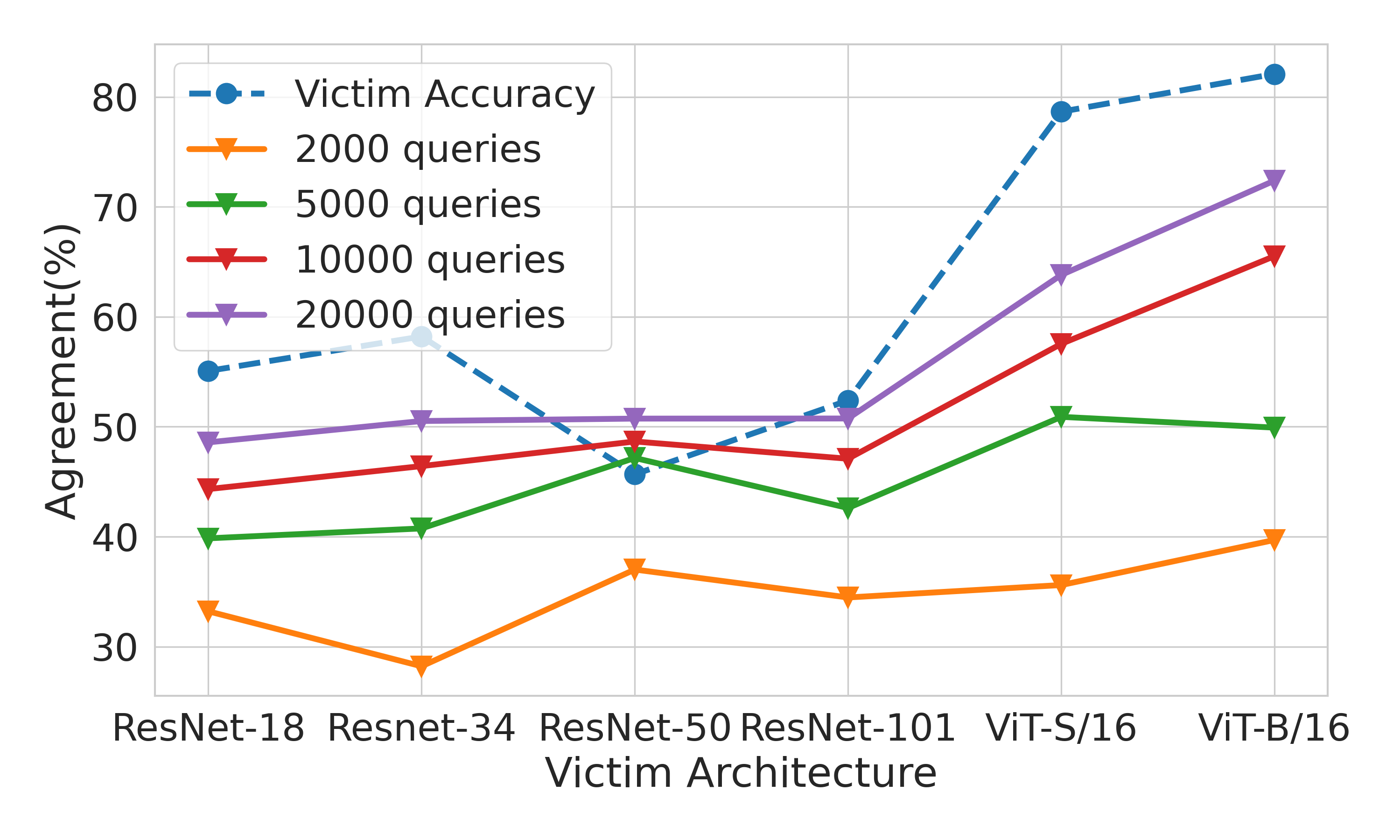}	
		\caption{Impact of varying the query budget, for `random' sample selection. Indoor-67 dataset, ViT-B/16 thief.}
		\label{fig:ablation1}
	\end{minipage}%
	\qquad
	\begin{minipage}{0.47\textwidth}
		\centering
		\includegraphics[width=\textwidth]{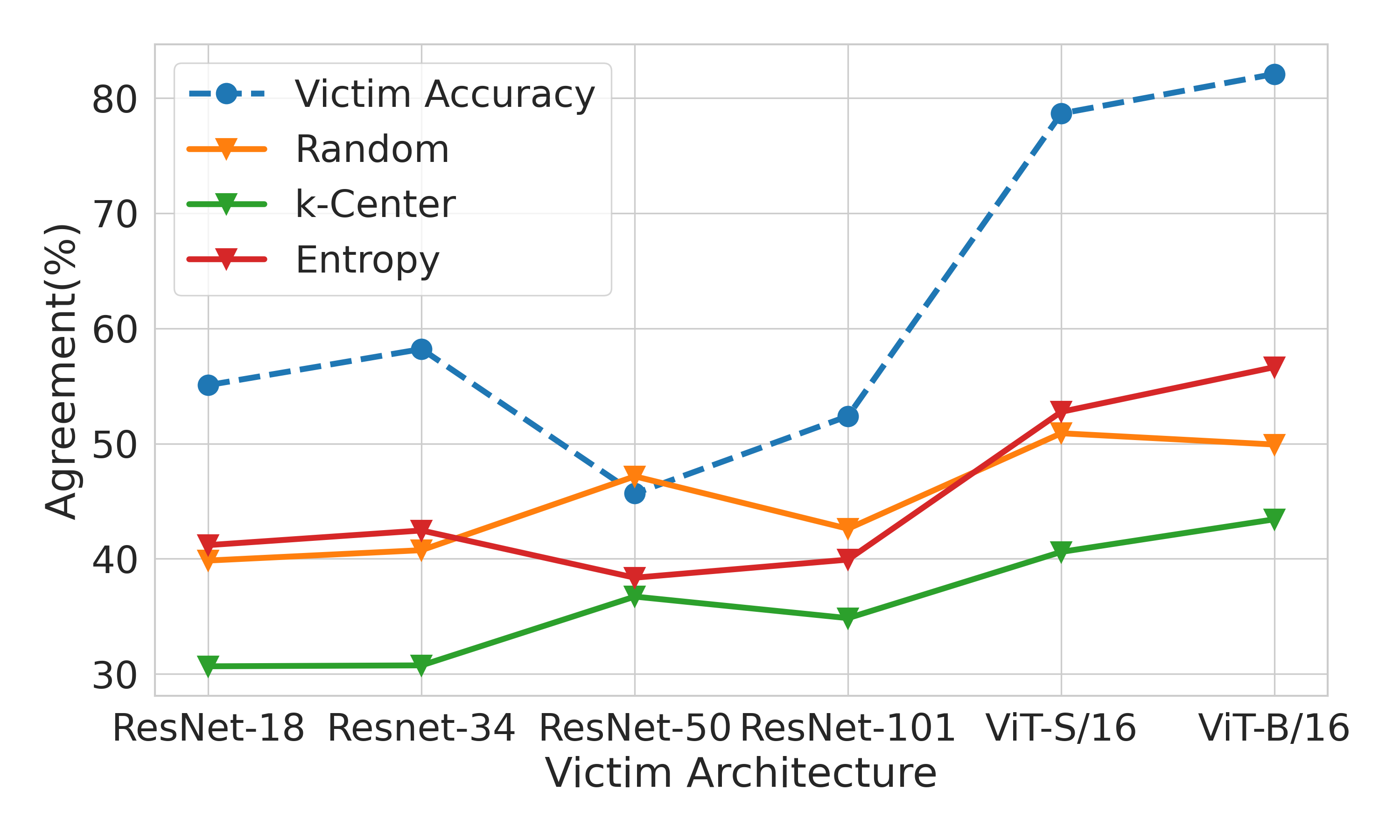}
		\caption{Impact of varying sample selection method, for a query budget of 5000. Indoor-67 dataset, ViT-B/16 thief.}
		\label{fig:ablation2}
	\end{minipage}
\end{figure}

\subsection{Qualitative Results}
We extend the qualitative results reported in \Cref{sec:tsne} for the CIFAR-10 dataset to other datasets. \Cref{fig:tsne_plots_indoor} and \Cref{fig:tsne_plots_caltech} show the t-SNE visualizations for the Indoor-67 and Caltech-256 datasets, respectively. The ViT backbones are able to form well-separated clusters on both the datasets, leading to better performance of the respective victim models, and also easier stealing.

\begin{figure*}[t]
	\centering
	\includegraphics[width=\linewidth]{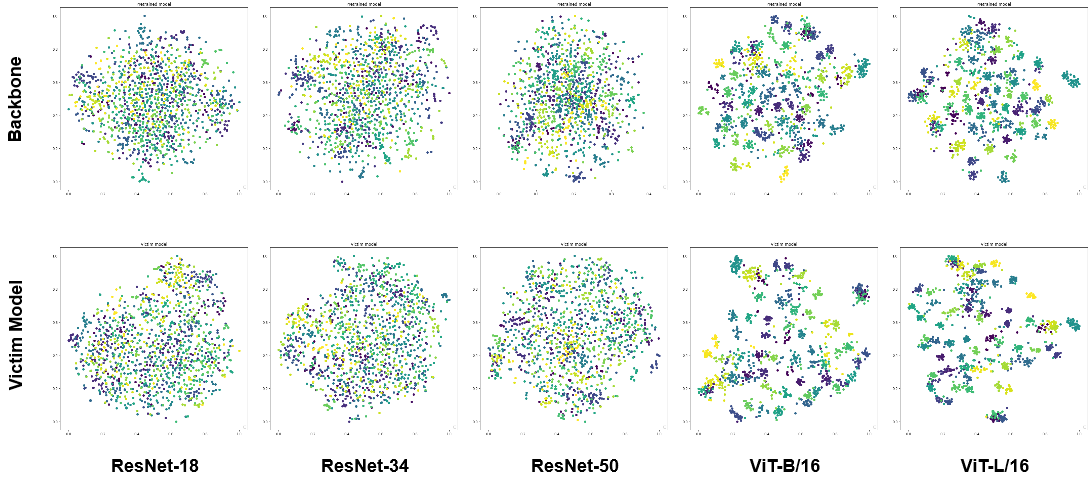}	
	\caption{t-SNE visualizations of embeddings for backbone models (top row), and corresponding victim models (bottom row) trained on Indoor-67 dataset, for various model architectures.}
	\label{fig:tsne_plots_indoor}
\end{figure*}

\begin{figure*}[t]
	\centering
	\includegraphics[width=\linewidth]{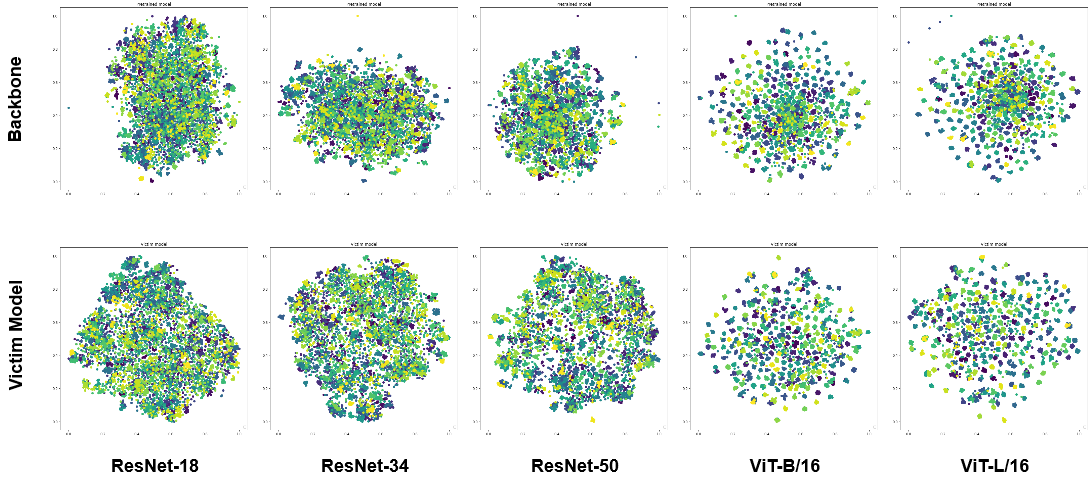}	
	\caption{t-SNE visualizations of embeddings for backbone models (top row), and corresponding victim models (bottom row) trained on Caltech-256 dataset, for various model architectures.}
	\label{fig:tsne_plots_caltech}
\end{figure*}

\end{document}